\documentclass[runningheads]{llncs}

\usepackage[preview]{eccv}

\usepackage{eccvabbrv}

\usepackage{graphicx}
\usepackage{booktabs}

\usepackage[accsupp]{axessibility}  

\usepackage{hyperref}

\usepackage{orcidlink}
\usepackage[table]{xcolor}
\newcommand{\method}{PAD\xspace}

\begin{document}

\title{Pose-Aware Diffusion for 3D Generation} 

\titlerunning{Pose-Aware Diffusion for 3D Generation}

\author{Zihan Zhou$^{*}$\inst{1}\and
Luxi Chen$^*$\inst{1} \and
Jingzhi Zhou\inst{2} \and Yuhao Wan\inst{2} \and Min Zhao\inst{3} \and Baoyu Fan\inst{4} \and Chongxuan Li$^\dagger$\inst{1}
}

\authorrunning{Zhou et al.}

\institute{Gaoling School of AI, Renmin University of China\and VCIP, School of Computer Science, Nankai University~\and THU-Bosch
MLCenter, Tsinghua University~\and Inspur Group 
}

\maketitle
  \let\oldthefootnote\thefootnote              
  \renewcommand{\thefootnote}{}
  \footnotetext{$^{*}$~Equal contribution.\quad $^{\dagger}$~Corresponding      
  author.}                                                                      
  \renewcommand{\thefootnote}{\oldthefootnote}    

\begin{abstract}
Generating pose-aligned 3D objects is challenging due to the spatial mismatches and transformation ambiguities inherent in decoupled canonical-then-rotate paradigms. To this end, we introduce Pose-Aware Diffusion (PAD), a novel end-to-end diffusion framework that synthesizes 3D geometry directly within the observation space. By unprojecting monocular depth into a partial point cloud and explicitly injecting it as a 3D geometric anchor, PAD abandons canonical assumptions to enforce rigorous spatial supervision. This native generation intrinsically resolves pose ambiguity, producing high-fidelity pose-aligned assets. Extensive experiments demonstrate that PAD achieves superior geometric alignment and image-to-3D correspondence compared to state-of-the-art methods. Additionally, PAD naturally extends to compositional 3D scene reconstruction via a simple union of independently generated objects, highlighting its robust ability to preserve precise spatial layouts. Code is available at  \url{https://github.com/ML-GSAI/PAD}.
  \keywords{3D Generation \and Diffusion Models \and Pose-Aware Generation}
\end{abstract}

\begin{figure}[!t]
  \centering
  \begin{subfigure}{\linewidth}
    \centering
    
    \includegraphics[width=1\linewidth]{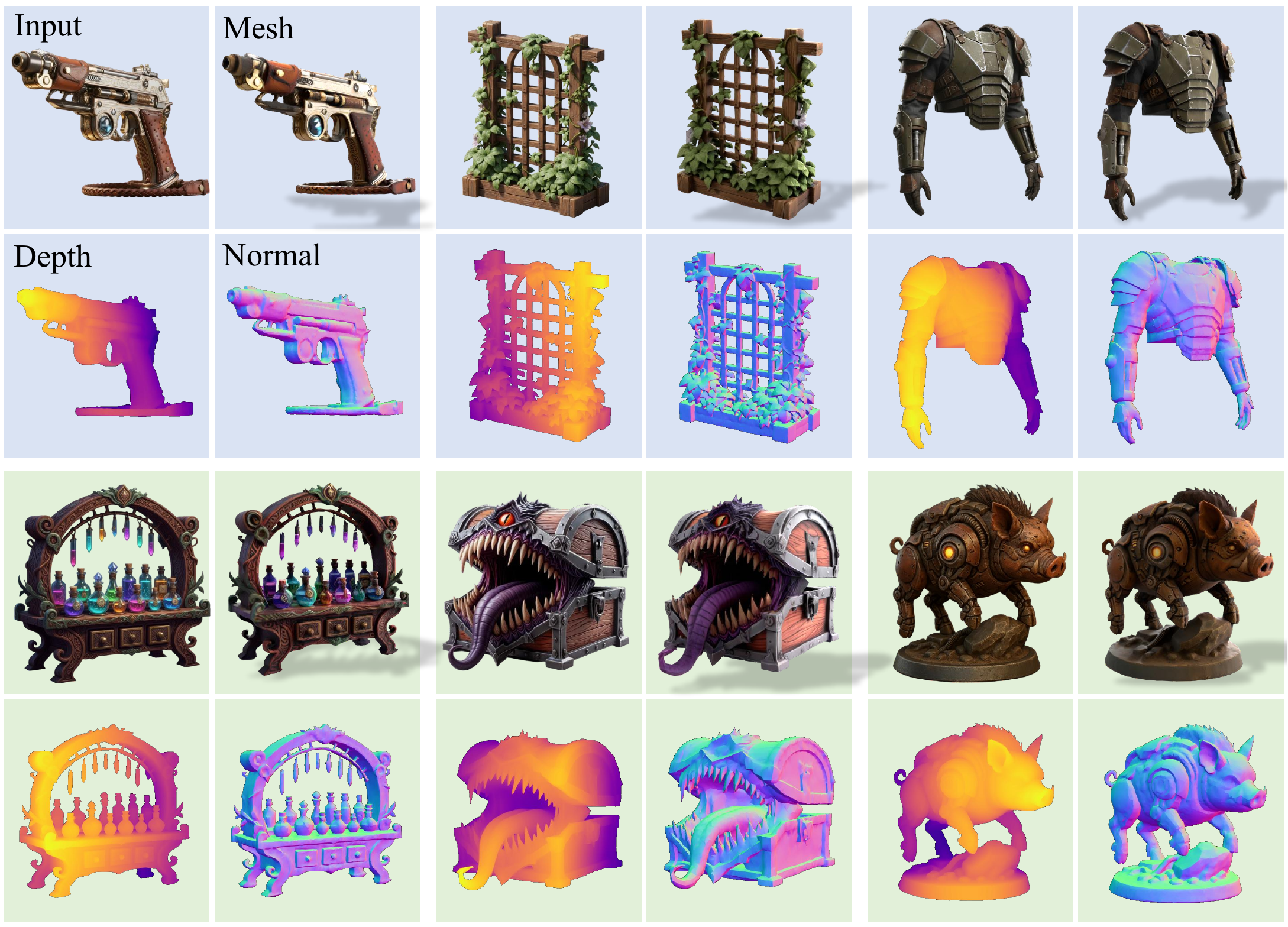}
    \caption{Results of high-fidelity pose-aligned object generation.}
    \label{fig:sub_top}
  \end{subfigure}
  \vspace{4mm} 
  \begin{subfigure}{\linewidth}
    \centering
    \includegraphics[width=\linewidth]{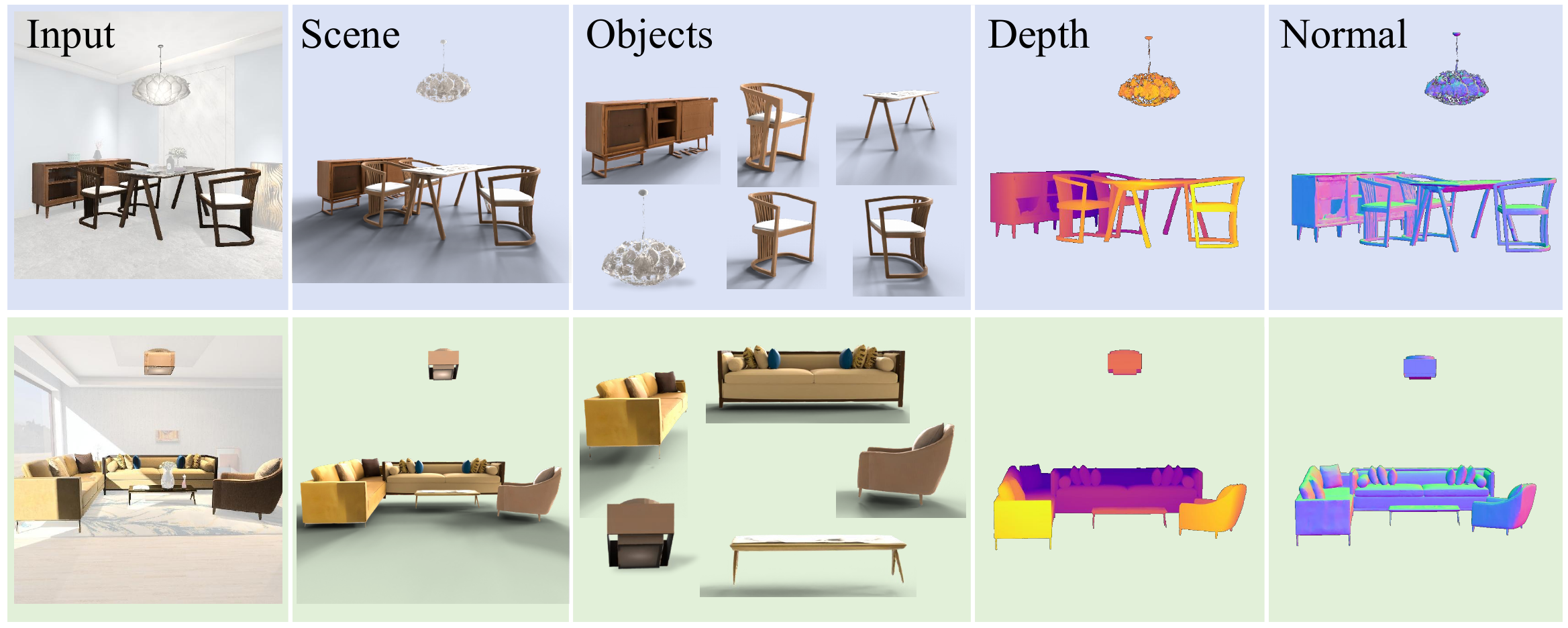}
    \caption{Results of compositional scene generation.}
    \label{fig:sub_bottom}
  \end{subfigure}
  \vspace{-1cm}
  \caption{\textbf{\method generates high-fidelity posed 3D objects and compositional scenes.}     (a) For individual objects, \method preserves intricate details and strict spatial alignment across challenging topologies. (b) By generating pose-aligned objects progressively, \method naturally extends to compositional scene generation while maintaining precise layout and geometric fidelity.}
  \label{fig:vertical_images}
  
\end{figure}

\section{Introduction}
\label{sec:intro}

Generating 3D assets from single images holds transformative potential for diverse applications, such as virtual reality, robotic simulation, and gaming. Recent feed-forward approaches and 3D diffusion models~\cite{hong2023lrm, tang2024lgm,xu2024grm,xu2024instantmesh,li2025triposg, xiang2025trellis2, zhang2024clay, xiang2025structured, hunyuan3d2025hunyuan3d} offer a principled solution by hallucinating plausible geometry and textures. However, ensuring precise spatial alignment between the generated 3D object and the input visual observation remains a formidable challenge. Most generative models operate in normalized, canonical coordinate spaces~\cite{shi2023zero123plus}, synthesizing isolated objects aligned to a fixed orientation, thereby breaking their spatial link to the physical world. Re-establishing this spatial link is essential for practical downstream applications, such as placing generated objects precisely back into a 3D scene. 

\begin{figure*}[t]
    \centering
    \includegraphics[width=1.0\linewidth]{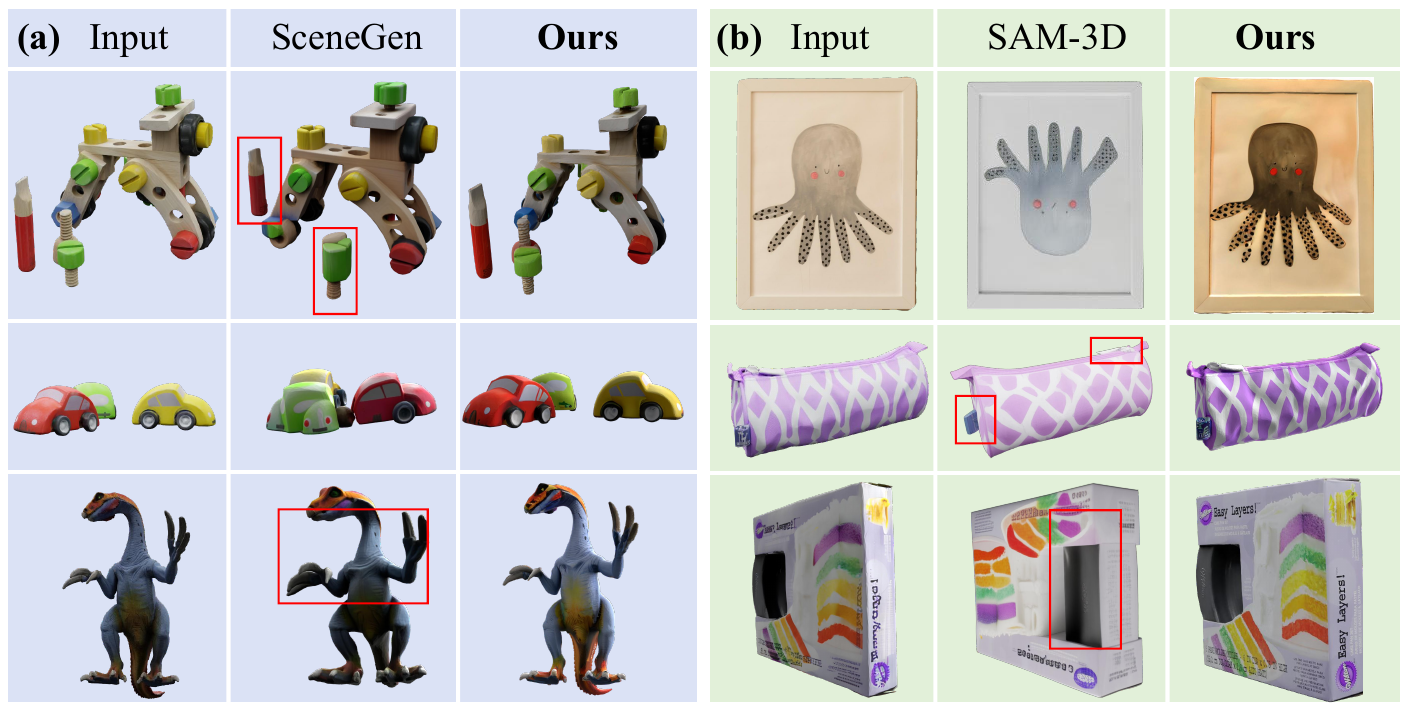} 
    \caption{\textbf{\method generates in the observation space to overcome the limitations of the canonical-then-rotate paradigm.} We present generation results from SceneGen~\cite{meng2025scenegen} (a) and SAM-3D~\cite{sam3dteam2025sam3d3dfyimages} (b), demonstrating spatial mismatch and pose-prediction failures. The red boxes indicate extraneous geometries or inverted rotations in the generated content. Our model ensures spatial alignment and avoids both issues.
    }  
    \label{fig:abla_intro}
\end{figure*}

To adapt canonical models for these tasks, current methods typically rely on a canonical-then-rotate paradigm~\cite{nie2020total3dunderstanding}, where an object is first generated in a canonical space and then rigidly transformed to match the observed pose with post-hoc optimization~\cite{zhao2025depr,zhou2024zero,han2025reparo} or concurrent pose estimation~\cite{sam3dteam2025sam3d3dfyimages,meng2025scenegen,huang2025cupid}. While these methods decouple generation and pose estimation to simplify training, they face notable limitations. First, because the generation process occurs within a canonical space while the input image is captured in the observation space, there can be a lack of direct connection to effectively transfer fine-grained details from the input. Consequently, even when the object undergoes a seemingly correct rotation, the generated shape may struggle to achieve pixel-level alignment with the input image (\cref{fig:abla_intro}a). Second, relying on a separate pose prediction step to map the canonical object back to the observation space introduces severe ambiguity, particularly for objects exhibiting symmetrical properties. This ambiguity increases the probability of transformation errors, such as inverted rotations (\cref{fig:abla_intro}b).

To this end, we propose Pose-Aware Diffusion (PAD) to achieve end-to-end, pose-aligned 3D generation directly within the observation space. We initialize \method using a pre-trained 3D model Hunyuan3D-2.1~\cite{hunyuan3d2025hunyuan3d}. To closely align the generated geometry with the input image, we extract a coarse 3D geometry (i.e., partial point cloud) using a metric depth estimator~\cite{wang2025moge} followed by unprojection. Then we inject this geometry into our network by concatenating it with the target latent along the sequence dimension of the diffusion transformer. By training this 3D conditioning model on our carefully curated paired data, we achieve end-to-end pose-aware generation. Abandoning the canonical space assumption, \method can yield high-fidelity, accurately pose-aligned objects directly without post-hoc pose estimation. Furthermore, it naturally extends to downstream compositional 3D scene generation tasks. By segmenting individual objects, extracting their respective partial point clouds, and generating each asset independently, a coherent 3D scene is ultimately achieved through a straightforward union without pose optimization per instance.

Our extensive experiments demonstrate \method's effectiveness in both high-quality object generation and downstream compositional scene generation tasks.  In particular, our method achieves superior image-3D correspondence and geometric alignment compared to state-of-the-art baselines~\cite{tang2023dreamgaussian,huang2025midi,meng2025scenegen,xu2024instantmesh,sam3dteam2025sam3d3dfyimages,siddiqui2026shaper} (see~\cref{tab:quant_obj}) while effectively resolving occlusions in the multi-object scene generation task (see~\cref{tab:quant_scene}).

In summary, our key contributions are:
\begin{itemize}
\item We propose \method, an end-to-end 3D generation framework that generates within the observation space, fundamentally eliminating pose ambiguity and spatial mismatch.
\item We introduce a direct 3D latent conditioning mechanism that encodes unprojected partial point clouds into the diffusion latent space, enforcing a strict spatial constraint for pose-aligned generation.
\item Extensive experiments demonstrate that \method not only achieves state-of-the-art performance in single-object generation and pose accuracy but also naturally enables high-quality compositional scene generation.
\end{itemize}

\section{Related works}
\label{sec:related_work}

\subsection{3D object generation}

Driven by the rapid advancement of powerful 2D image synthesis models~\cite{rombach2022high,podell2023sdxl}, pioneering 3D generative frameworks~\cite{poole2022dreamfusion,wang2024prolificdreamer,chen2024microdreamer,liang2023luciddreamer, tang2023dreamgaussian} have extensively utilized optimization methods based on differentiable rendering~\cite{mildenhall2021nerf, kerbl20233d, muller2022instant} to lift 2D generative priors into high-fidelity 3D representations. Subsequent research has transitioned toward multi-view diffusion~\cite{shi2023zero123plus, wang2023imagedream, shi2023mvdream} or video generative models~\cite{voleti2024sv3d,chen2024v3d} to ensure spatial consistency. Concurrently, there is a paradigm shift from iterative optimization~\cite{long2024wonder3d, liu2023syncdreamer} to feed-forward large reconstruction models~\cite{hong2023lrm, tang2024lgm,xu2024grm,xu2024instantmesh}, which directly generate 3D assets in a single pass. Current state-of-the-art frameworks have adopted native 3D generative modeling \cite{li2025triposg, xiang2025trellis2, zhang2024clay, xiang2025structured, hunyuan3d2025hunyuan3d}, which enables scaling on massive 3D datasets~\cite{deitke2023objaverse,deitke2023objaversexl,khanna2024habitat,collins2022abo} to directly map text or images to 3D representations without relying on intermediate multi-view or video. While this paradigm significantly elevates generation fidelity, these models predominantly operate within predefined canonical coordinate spaces. Consequently, directly generating a 3D asset that strictly adheres to the unposed, arbitrary perspective of a specific input image remains a critical challenge. \method addresses these limitations by explicitly anchoring the generated object's pose with the input image, enabling end-to-end generation in the observation space without the need for additional post-optimization or manual viewpoint alignment.

\subsection{3D scene generation and composition}

Generating 3D scenes from a single or sparse set of images remains a challenging task. One group of methods adopts an explore-and-inpaint strategy~\cite{yu2024wonderworld, yu2024wonderjourney, lucid}. These approaches typically utilize visual perceptual models like depth estimators~\cite{bhat2023zoedepth, wang2025moge} or dense stereo models~\cite{wang2024dust3r, wang2025vggt} to warp input images into novel viewpoints, followed by 2D image or video diffusion models~\cite{xu2024camco, wang2024motionctrl} to inpaint missing regions and incorporate them into 3D structure through optimization~\cite{chen2025flexworld, wan2025geoworld, seed2025seed3d} or reconstruction models~\cite{liang2024wonderland, zhai2025stargen, viewcrafter}. While such methods can produce visually plausible results from training views, their reliance on 2D diffusion priors lacking explicit 3D awareness often leads to error accumulation during iterative viewpoint warping, resulting in geometric inconsistencies and degraded quality in unseen regions. Alternatively, compositional scene generation approaches~\cite{zhou2024zero,ardelean2025gen3dsr,zhao2025depr,huang2025midi,meng2025scenegen,sam3dteam2025sam3d3dfyimages,han2025reparo,yao2025cast} decompose the scene into individual objects, which are synthesized separately. However, these methods require precise estimation and delicate adjustment of each object's pose to integrate them into the original scene, often leading to misaligned spatial relationships. To overcome both the geometric distortions of incremental inpainting and the spatial misalignments of traditional compositional assembly, \method leverages metric depth estimation models to extract a 3D structural scaffold. Based on this global layout, our pose-aware diffusion model synthesizes each scene component directly. This approach natively bypasses post-hoc pose optimization, ensuring that the generated scene maintains spatial fidelity and superior geometric consistency with the input observation.

\section{Method}

\begin{figure*}[t]
    \centering
    \includegraphics[width=1.0\linewidth]{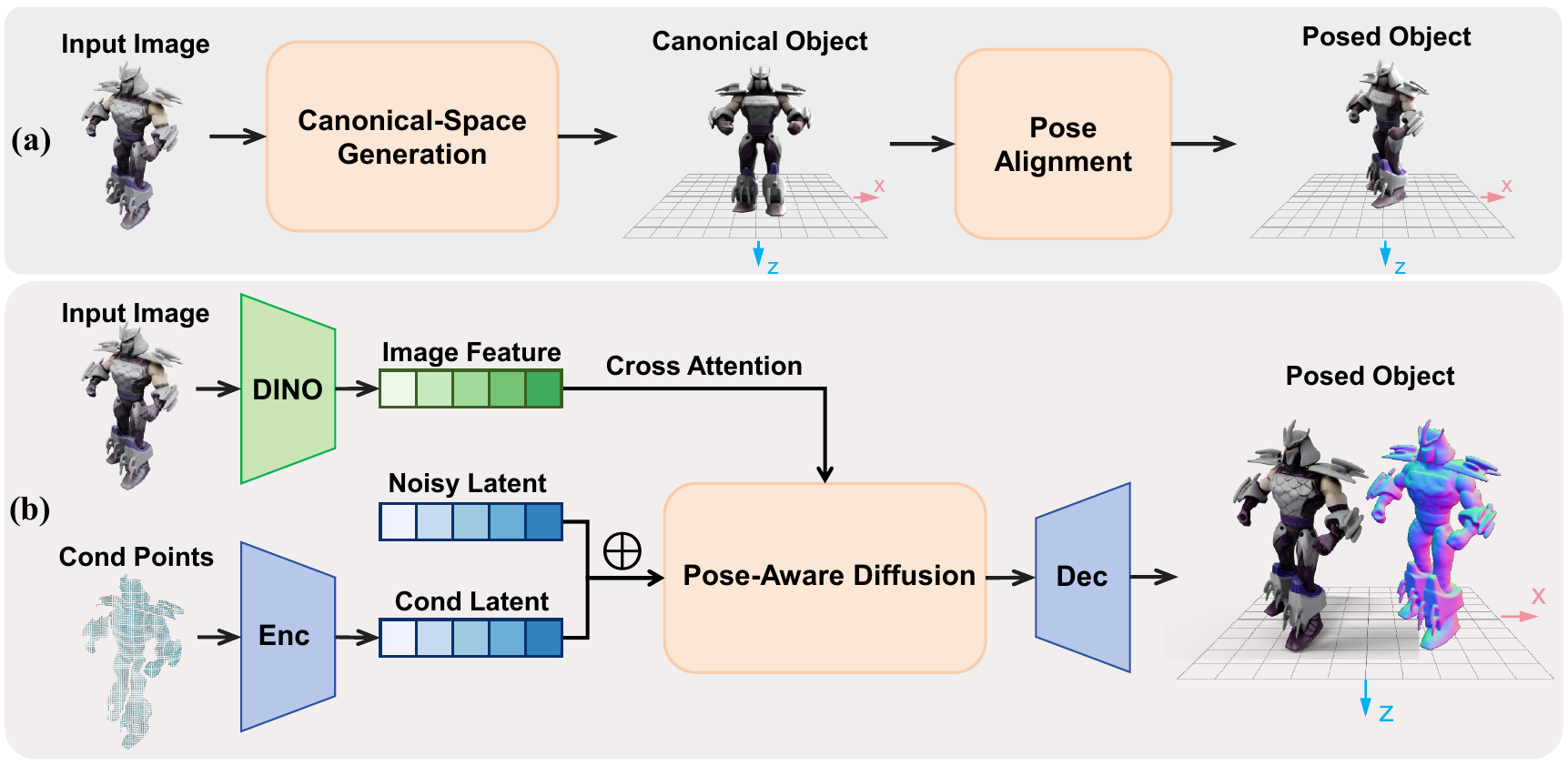} 
    \caption{\textbf{Overview of \method.}  
    (a) Existing decoupled paradigms typically follow a canonical-then-rotate approach, generating an object in a canonical space, followed by pose alignment. (b) In contrast, \method synthesizes complete 3D meshes directly in the observation space. The model concatenates a geometric latent encoded from the visible partial condition point cloud. From a single image, \method yields a pose-aligned 3D object directly, eliminating the need for post-hoc pose registration.
    }  
    \label{fig:method}
\end{figure*}

In this section, we first introduce the preliminaries for \method. Subsequently, we present our core pose-aware diffusion model for end-to-end, pose-aligned 3D object generation in Sec.~\ref{sec:objgen}. Finally, we describe how this observation-space generation paradigm naturally extends to compositional scene generation in Sec.~\ref{sec:scenegen}.

\subsection{Preliminaries}
\label{sec:preliminary}

\subsubsection{Vecset VAE.}

To compress the 3D point cloud into a compact latent space, we adopt a vecset VAE, an architecture widely utilized in 3D generative models~\cite{zhang20233dshape2vecset, hunyuan3d2025hunyuan3d, zhang2024clay} due to its flexibility.
Let $\mathcal{E}$ and $\mathcal{D}$ denote the encoder and decoder, respectively. Given a 3D object $x$, the encoder extracts a sequence of latent tokens $z = \mathcal{E}(x)$, which serves as the input for the diffusion model. Conversely, the decoder reconstructs the 3D geometry from the generated latents: $\hat{x} = \mathcal{D}(z)$. The output of the decoder is formulated as a signed distance field (SDF). Specifically, for any queried spatial coordinates $p \in \mathbb{R}^3$, the decoder predicts its corresponding SDF value: $s = \mathcal{D}(z, p)$. This implicit representation enables continuous geometry querying at arbitrary resolutions. To obtain the final explicit 3D mesh, the marching cubes algorithm~\cite{lorensen1998marching} is applied to extract the zero-level set from the predicted SDF grid.

A key architectural feature of vecset VAE is the permutation invariance of its latent space. The latent $z$ is represented as a set of tokens. Because the decoder employs symmetric aggregation and global attention mechanisms, permuting the token sequence in $z$ leaves the predicted SDF value at any spatial point $p$ unchanged. This property ensures that the latent space models the underlying geometry as an unordered set of structural features, offering enhanced flexibility and scalability for diffusion-based generation.

\subsubsection{Native 3D diffusion model.}
We build our generative backbone upon the flow matching framework~\cite{lipman2022flow,hunyuan3d2025hunyuan3d}.
Specifically, let $z_1 \in \mathbb{R}^{L \times C}$ denote the latent representation of a 3D object extracted by the VAE, and $z_0 \sim \mathcal{N}(0, I)$ denote the initial noise. The forward process constructs an interpolated state $z_t = t z_1 + (1-t) z_0$, where $t \in [0, 1]$ represents the timestep. The model acts as a velocity field estimator $v_\theta(z_t, t, c)$, parameterized by $\theta$ and conditioned on $c$. The training objective of the model is:

\begin{align}
    \min_{\theta} \mathbb{E}_{t\sim\mathcal{U}(0,1), z_0\sim \mathcal{N}(0, {I})}\left[\|v_{\theta}({z}_t, t, {c})-(z_1-z_0)\|_2^2\right]. \label{eq:diff-loss}
\end{align}

In our work, we leverage a pre-trained DiT-based~\cite{peebles2023scalable, hunyuan3d2025hunyuan3d} velocity estimator and focus on adapting the conditioning mechanism $c$ for pose-aware generation.

\subsection{Pose-aware diffusion model}
\label{sec:objgen}

To overcome the limitations of canonical-space generation and the resulting pose ambiguities (as discussed in~\cref{sec:intro}), we propose the pose-aware diffusion (\method). \method addresses this challenge by formulating 3D object generation as an end-to-end process within the observation space. By conditioning the model on partial point clouds inferred from the input image, we create a direct pixel-level link between the input and the generated 3D. This allows for end-to-end generation without auxiliary pose prediction, ensuring the synthesized mesh maintains high structural and detail fidelity relative to the input image.

\subsubsection{End-to-end generation in observation space.}

Given the input image $I$, existing decoupled generation paradigms first generate a 3D mesh $M$ in canonical space, and subsequently estimate a rigid pose $T$. The final posed object is then obtained via a transform $\hat{M} = T \circ M$ (~\cref{fig:method}a). Because the canonical geometry distribution $p(M | I)$ and the pose prediction $p(T | I)$ are decoupled, any estimation error in either step will lead to severe spatial misalignment.

To resolve this, \method unifies generation and alignment into an end-to-end framework operating directly in the observation space (~\cref{fig:method}b). Rather than relying on canonical assumptions, we explicitly model the $1$-to-$1$ mapping $p(\hat{M} | I)$ using~\cref{eq:diff-loss}. Consequently, the generated geometry is locked to the input viewpoint, eliminating the need for separate pose estimation or post-hoc pose registration.

\subsubsection{Latent point conditioning.}
To create a pixel-level link between the input and the generated 3D, we introduce explicit spatial constraints rather than relying solely on 2D image features. We utilized a pretrained depth estimator~\cite{wang2025moge} to derive spatial information from the input image, which represents the visible surface of the object in 3D space.

While a naive approach might use this estimated depthmap for cross-attention conditioning, such image-space operations often critically degrade high-frequency geometric precision (see the ablation in Fig.~\ref{fig:abla}b). Therefore, we propose a direct 3D latent conditioning strategy. We first unproject the depthmap to partial point cloud $P_{\text{partial}}$. Then we utilize the pre-trained vecset VAE encoder (introduced in Sec.~\ref{sec:preliminary}) to map the partial point cloud into a compact geometric latent space: $z_{\text{geo}} = \mathcal{E}(P_{\text{partial}})$. Before encoding, the point cloud is padded or cropped to a fixed size.

Crucially, this geometric latent feature $z_{\text{geo}}$ is spatially aligned with the target diffusion latent space. During the flow matching denoising process, we directly concatenate $z_{\text{geo}}$ with the interpolated latent $z_t$ along the sequence dimension, as shown in~\cref{fig:method}. The velocity estimator network is formulated as:
\begin{equation}
v_\theta(z_t, t, c) := v_\theta(\text{Concat}(z_t, z_{\text{geo}}), t, c_{\text{img}}),
\end{equation}
where $c_{\text{img}}$ represents global semantic conditioning derived from the input image. This explicit concatenation acts as a strong geometric anchor, forcing the network to physically grow the complete watertight mesh outward from the encoded visible surface, ensuring seamless alignment with the initial observation.

\subsubsection{Stabilizing Pose-Aware Training.}
To leverage the strong geometric priors of large-scale 3D pretrained models, we initialize our network with Hunyuan3D-2.1~\cite{hunyuan3d2025hunyuan3d}. However, fine-tuning it into a pose-aware 3D diffusion model introduces significant stability challenges due to the domain shift from the canonical space to the observation space. Empirically, we observe that incorporating point cloud conditioning also effectively mitigates this instability (see ablation study in Fig.~\ref{fig:abla}a,d). Specifically, we condition the generation process on ground-truth partial point clouds captured directly within the observation space. To this end, we curate high-quality datasets from Objaverse~\cite{deitke2023objaverse} and 3D-Front~\cite{fu20213d}, filtering out geometrically trivial meshes, and rendering 24 views per object, along with their depth maps. Crucially, to provide precise spatial supervision, the ground-truth meshes are explicitly rotated to align perfectly with each rendered view. We then back-project the ground-truth depth maps to construct the exact visible partial point cloud $P_{\text{partial}}$, serving as a strong 3D geometric anchor.

While this stabilizes the training, the model becomes overly reliant on the partial point cloud. Consequently, it suffers from severe performance degradation during inference when exposed to the inevitable noise produced by depth estimators. To break this dependency, we deliberately corrupt the conditioning data by randomly replacing the ground truth with estimated depth and injecting noise into the partial point clouds. This depth noise augmentation prevents the model from degenerating into a trivial autoencoder, forcing it to robustly hallucinate missing geometry. As shown in our ablation study (Fig.~\ref{fig:abla}c,d), this approach significantly improves the model's generalization capability and robustness against imperfect partial point clouds during inference. See Appendix A for more details.

\subsection{Compositional 3D scene generation}
\label{sec:scenegen}

A natural extension of our pose-aware generation paradigm is the compositional generation of multi-object scenes. Because \method natively synthesizes geometries with their poses aligned with the camera frame, we can seamlessly generate entire 3D scenes by assembling independently generated components. This compositional approach fundamentally circumvents the need for multi-view joint optimization or post-hoc pose estimation in traditional scene-generation pipelines.

To process a single scene image $I_{\text{scene}}$, we first establish a global metric coordinate system. We employ a metric monocular depth estimator (e.g., MoGe~\cite{wang2025moge}) to predict a depth map $D$ and its corresponding camera intrinsics $K$. By unprojecting the pixels into 3D space, we obtain a global, partial scene point cloud $P_{\text{scene}}$:
\begin{equation}
    P_{\text{scene}} = \{ K^{-1} [u, v, 1]^T \cdot D(u,v) \mid (u,v) \in I_{\text{scene}} \}.
\end{equation}

This partial point cloud explicitly encodes the global spatial layout, including the relative positions and metric scales of the visible scene elements. To process the scene compositionally, we segment the image into individual objects. For the $i$-th object, we extract its partial frontal point cloud $P^{(i)}_{\text{partial}}$ and the corresponding cropped image patch $I^{(i)}$.

We independently process each tuple $(I^{(i)}, P_{\text{partial}}^{(i)})$ through our pose-aware diffusion model. Crucially, because the geometric condition $P_{\text{partial}}^{(i)}$ is anchored to its original coordinates within the global camera frame, the model synthesizes the completed watertight mesh $M^{(i)}$ directly in its correct rotation and scale. The final coherent 3D scene is achieved through a straightforward union of all generated individual assets: $M_{\text{scene}} = \bigcup_{i} M^{(i)}$. 

In contrast to representative scene generation methods based on post-hoc optimization pipelines~\cite{zhao2025depr,zhou2024zero}, which require pose registration per object, our approach naturally bypasses these spatial misalignments. By reducing scene generation to a set of independent, spatially anchored object completion tasks, \method inherently resolves mutual occlusions and preserves strict geometric consistency.

\section{Experiments}
\label{sec:experiments}
We present the implementation details for \method, comparison on posed object generation and compositional scene generation, and ablation studies sequentially.

\subsection{Implementation details}
We finetuned our pose-aware diffusion model based on Hunyuan3D-2.1~\cite{hunyuan3d2025hunyuan3d}. During training, we use the AdamW optimizer with a cosine learning rate scheduler with a peak learning rate of $1 \times 10^{-5}$, and set the batch size to 16 on 8 NVIDIA A100 GPUs. The training process spans 50k iterations, which takes about 2 days. We employ gradient clipping with a threshold of 1.0 to stabilize the training. For conditioning, we use a classifier-free guidance (CFG) drop rate of 0.1 and randomly downsample or duplicate the 
partial point clouds to 81920, which matches the input specification of our shape encoder. For inference, we keep the original settings unchanged and use 50 steps with a CFG strength of 3.0 to generate the final latent. See Appendix A for more details.

\subsection{Comparison on posed object generation}
To validate the effectiveness of our pose-aware generation strategy, we conduct a comprehensive quantitative comparison against state-of-the-art 3D generation methods. The evaluation is conducted at the object level using RGBA rendered images as conditional inputs, and the depth is estimated by MoGe~\cite{wang2025moge} to simulate realistic inference scenarios where only a single view and estimated depth are available.

\subsubsection{Evaluation datasets.} Following prior works~\cite{meng2025scenegen, huang2025cupid}, we perform our evaluation on a rigorous test set derived from the Google Scanned Object (GSO)~\cite{downs2022google} dataset. The dataset contains about 1,000 objects that were strictly excluded from the training phase. For each test object, we render a single RGBA image from a random viewpoint to serve as the input condition. Crucially, the ground truth mesh is transformed into the camera coordinate system of the input view. This setup allows us to assess the generated object's alignment with the input observation without the need for additional registration steps.

\subsubsection{Evaluation metrics.}
We employ a multi-dimensional metric suite to assess geometric accuracy, pose alignment, and semantic fidelity. Following prior works~\cite{huang2025midi, meng2025scenegen}, we use the Chamfer distance (CD), Volumetric Intersection over Union of bounding boxes (IoU-B), and F-Score with a threshold of 0.005 to evaluate the generated object and the pose-aware ground truth objects. To further validate the generated quality, we also use pretrained 3D models like ULIP~\cite{xue2023ulip} and Uni3D~\cite{zhou2023uni3d} to calculate the matching score between input images and generated objects following prior research~\cite {hunyuan3d2025hunyuan3d, seed2025seed3d, hunyuan3d2026hy3d}.

\subsubsection{Evaluation baselines.}
We select open-source baselines that represent distinct paradigms in the field. These include DreamGaussian~\cite{tang2023dreamgaussian} as a representative SDS-based method with front-view optimization and InstantMesh~\cite{xu2024instantmesh} as a large-scale feed-forward reconstruction model. Furthermore, we compare against MIDI~\cite{huang2025midi}, SceneGen~\cite{meng2025scenegen}, SAM-3D~\cite{sam3dteam2025sam3d3dfyimages}, and ShapeR~\cite{siddiqui2026shaper} to evaluate performance against methods specifically designed for scene-level or layout-conditioned generation.

\subsubsection{Qualitative and quantitative comparisons.}
Our quantitative results are presented in \cref{tab:quant_obj}. \method demonstrates superior performance in generating high-fidelity 3D assets that closely align with the input images. Specifically, our approach achieves the highest F-Score and image-to-3D similarity metrics, alongside the lowest Chamfer distance across the evaluation set. These results indicate that our generated models faithfully preserve the semantic and visual characteristics of the reference conditions while maintaining geometrically precise 3D structures.

\begin{figure*}[t]
    \centering
    \includegraphics[width=1.0\linewidth]{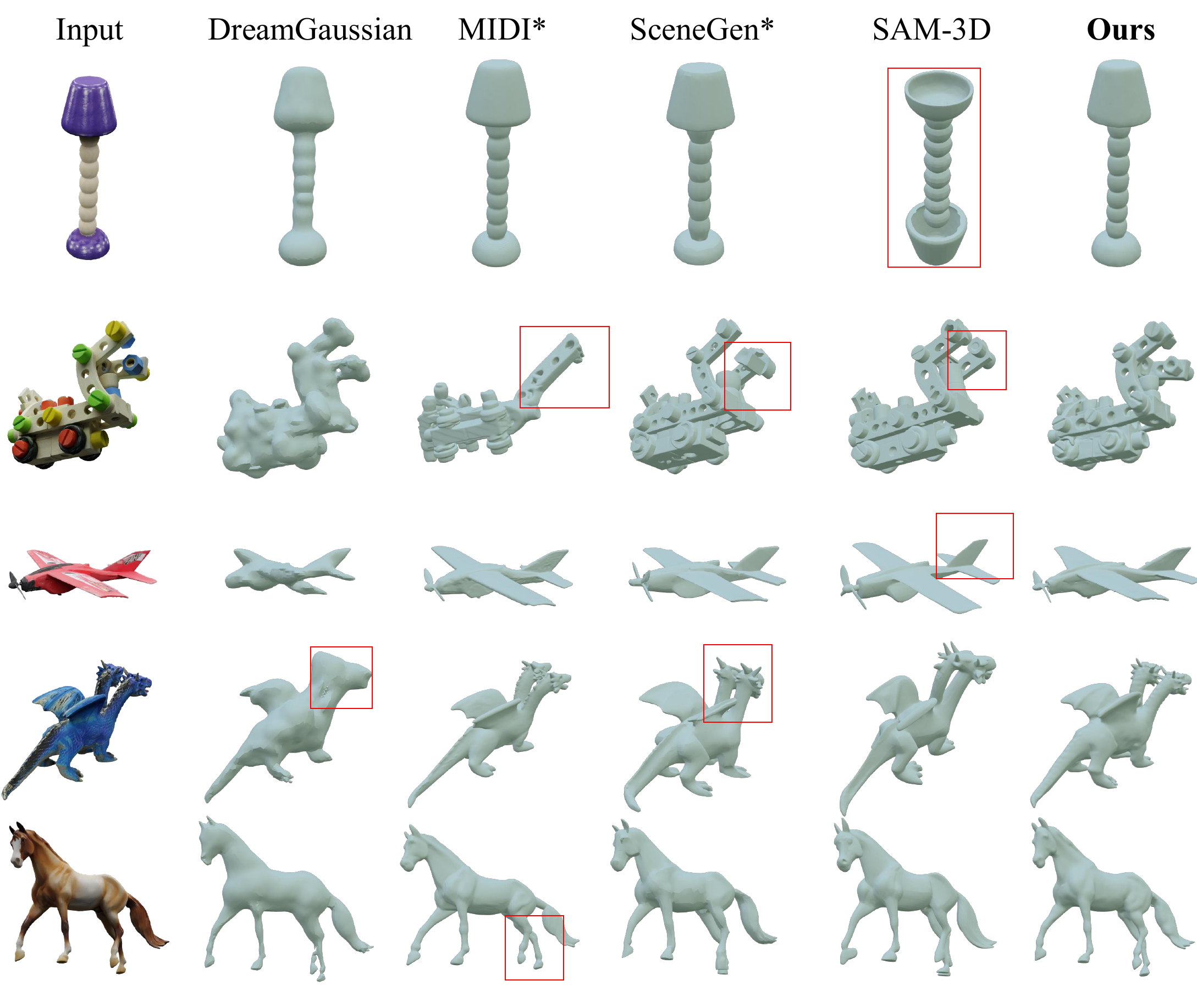} 
    \caption{\textbf{Qualitative comparison on posed object generation.} \method accurately generates 3D objects with strict spatial alignment from single images. While baseline methods frequently produce misaligned geometry or structural distortions (highlighted by red boxes), our approach robustly preserves intricate geometric details and the exact pose of the input objects. Notably, MIDI~\cite{huang2025midi} and SceneGen~\cite{meng2025scenegen} are rotated to align with the input using FoundationPose~\cite{wen2024foundationpose} with ground truth depth.
    }
    \label{fig:qual-comp-posed-obj}
\end{figure*}

We provide qualitative comparisons in \cref{fig:qual-comp-posed-obj} to illustrate the superiority of our approach further. Existing methods exhibit noticeable drawbacks: SAM-3D often produces misaligned outputs that lack the intricate visual details present in the reference images. Conversely, DreamGaussian maintains better pose alignment but is fundamentally limited by its rough and low-quality 3D geometry. Moreover, generating assets in a unified canonical space remains a challenge for MIDI and SceneGen, which produce inconsistent global orientations that severely complicate later pose recovery stages. \method explicitly addresses these bottlenecks. By jointly optimizing for image alignment and geometric precision, our model yields high-fidelity, pose-aligned 3D assets with rich structural and geometric details.
\newcolumntype{C}[1]{>{\centering\arraybackslash}p{#1}}
\begin{table}[t]
  \caption{\textbf{Quantitative comparision on posed object generation}. \method demonstrates superior geometric quality and precise spatial alignment compared to baseline methods. The $^\star$ means the generated model is rotated to align with the input image using FoundationPose~\cite{wen2024foundationpose} using ground truth depth.
  }
  \label{tab:quant_obj}
  \centering
  \begin{tabular}{lccccc}
    \toprule
    Metric & $\,$CD($\times10^{-3}$)$\downarrow\,$ & $\,$F-Score$\uparrow\,$ & $\,$IoU-B$\uparrow\,$ & $\,$Uni3D$\uparrow\,$ & $\,$ULIP$\uparrow\,$\\
    \midrule
    
    DreamGaussian~\cite{tang2023dreamgaussian} & 110.19 & 0.070 & 0.701 & 0.199 & 0.134 \\
    InstantMesh~\cite{xu2024instantmesh}   & 167.05 & 0.052 & 0.619 & 0.257 & 0.171 \\
    MIDI$^\star$~\cite{huang2025midi}   & 65.10  & 0.167 & 0.813 & 0.272 & 0.204  \\
    SceneGen$^\star$~\cite{meng2025scenegen}  & 57.06  & 0.172 & 0.838 & 0.274 & \textbf{0.218} \\
    
    ShapeR~\cite{siddiqui2026shaper}     & 230.1 & 0.052 & 0.448 & 0.148 & 0.133 \\
    
    SAM-3D~\cite{sam3dteam2025sam3d3dfyimages}     & 56.77 & 0.188 & 0.802 & 0.263 &  0.212 \\
    \cellcolor{gray!10}\textbf{Ours}     &\cellcolor{gray!10} \textbf{48.76} & \cellcolor{gray!10}\textbf{0.204} & \cellcolor{gray!10}\textbf{0.863} & \cellcolor{gray!10}\textbf{0.275} & \cellcolor{gray!10}0.213 \\
    
  \bottomrule
  \end{tabular}
\end{table}

\begin{table}[t]
  \caption{\textbf{Quantitative comparision on composistional scene generation}. \method shows superior geometrical quality compared with baseline methods. 
  }
  \label{tab:quant_scene}
  \centering
  \begin{tabular}{lccc}
    \toprule
    Metric & $\,$CD($\times10^{-3}$)$\downarrow\,$ & $\,$F-Score$\uparrow\,$ & $\,$IoU-B$\uparrow\,$ \\
    \midrule
    MIDI~\cite{huang2025midi}   & 169.71 & 0.069 & 0.626 \\
    SceneGen~\cite{meng2025scenegen}  & 177.11 & 0.078 & 0.635  \\
    SAM-3D~\cite{sam3dteam2025sam3d3dfyimages}     & 61.02 & 0.196 & 0.816  \\
    \cellcolor{gray!10}\textbf{Ours}     & \cellcolor{gray!10}\textbf{56.24} & \cellcolor{gray!10}\textbf{0.212} & \cellcolor{gray!10}\textbf{0.833} \\
  \bottomrule
  \end{tabular}
\end{table}

\begin{figure*}[t]
    \centering
    \includegraphics[width=1.0\linewidth]{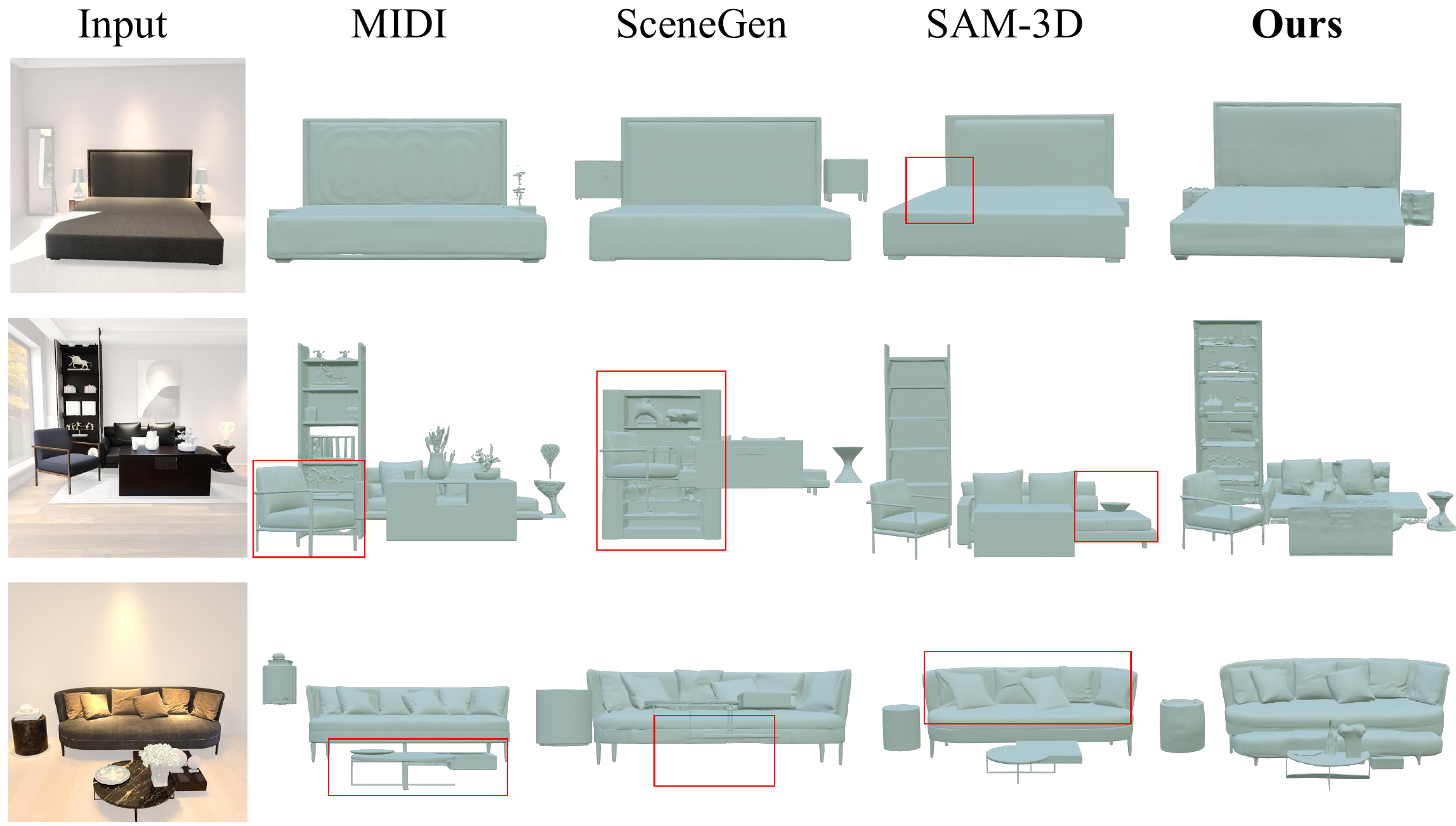} 
    \caption{\textbf{Qualitative comparison on compositional scene generation.} \method accurately generates complex 3D scenes from single input images. While baseline methods frequently exhibit geometric distortions, missing object parts, or spatial misalignments (highlighted by red boxes), our approach recovers detailed individual assets and maintains a faithful compositional layout.}  
    \label{fig:qual-comp-scene}
\end{figure*}

\begin{figure*}[t]
    \centering
    \includegraphics[width=1.0\linewidth]{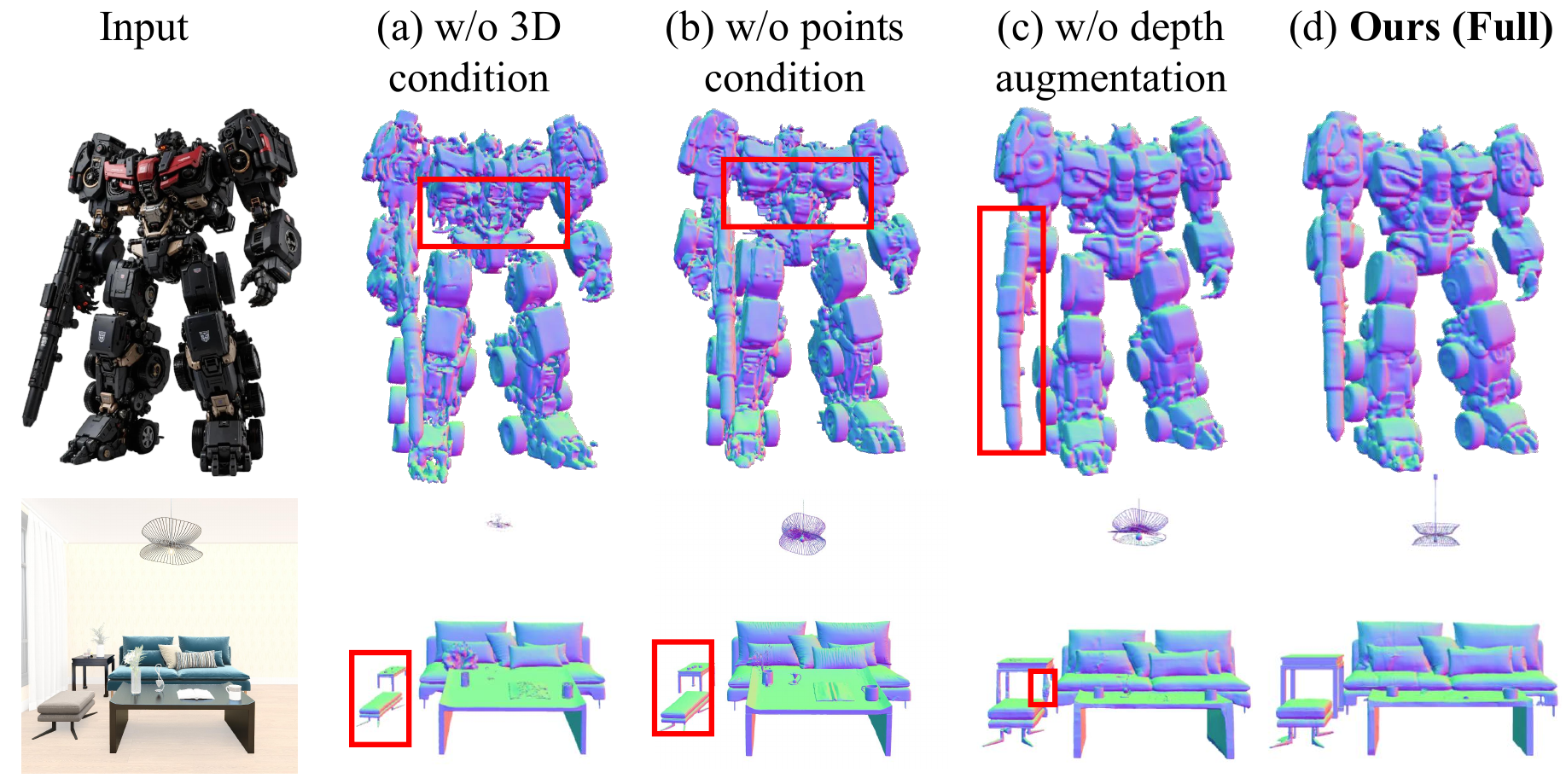} 
    \caption{\textbf{Ablation study.} We evaluate the key components of \method. Removing the 3D condition (a) or the point-based representation (b) leads to severe structural distortions and spatial misalignments. Omitting depth noise augmentation (c) causes collapsed surfaces and artifacts when handling imperfect depth inputs. In contrast, our full model (d) robustly generates high-fidelity, perfectly aligned geometry. Red boxes highlight the degraded regions in the ablated models. A slight rotation is applied to demonstrate the results better.}  
    \label{fig:abla}
\end{figure*}

\subsection{Comparison on compositional scene generation}

Since \method can accept pose-aware point cloud inputs and generate pose-aligned objects that match partial point clouds, it naturally extends to compositional scene generation.  Following the protocols in prior work~\cite {huang2025midi,meng2025scenegen}, we use 3D-FUTURE~\cite{fu20213dfuture} as the evaluation dataset. We select the first 200 scenes in the test set for evaluation. To further enhance the visual quality of the model, we perform additional fine-tuning on the 3D-FUTURE training set. We adopt standard metrics from prior work, such as Chamfer Distance and F-score. As Uni3D and ULIP are commonly used to evaluate objects, we do not include them in the evaluation matrices. We select MIDI~\cite{huang2025midi}, SceneGen~\cite{meng2025scenegen}, and SAM-3D~\cite{sam3dteam2025sam3d3dfyimages} as our baselines.

We present the qualitative results in \cref{fig:qual-comp-scene} and the quantitative results in \cref{tab:quant_scene}. All methods can generate compositional scenes based on a single scene image and object masks, and they produce reasonable geometry and relative spatial relationships. However, MIDI tends to over-hallucinate object geometry under the masks; SceneGen often produces floating objects with poor spatial relationships, and SAM-3D exhibits geometric details inconsistent with the input image. In contrast, our results demonstrate superior spatial positioning and output geometric structures that align faithfully with the reference images.

\subsection{Ablation Study}
\label{sec:abla}
In this section, we analyze the key components of our framework to verify their individual contributions.

\subsubsection{Effectiveness of latent point conditioning.}
As shown in Fig.~\ref{fig:abla}a, removing the conditioning with only image conditions in \method leads to significant structural distortions and misalignment with the input view. Without explicit geometric information, the diffusion process fails to localize objects' relative positions accurately and produces noisy meshes with degraded surface smoothness. This confirms that latent point conditioning is essential for maintaining the spatial correspondence between the 2D input and the generated 3D geometry.

\subsubsection{Lifting depth maps to 3D point clouds.}
We validate our point conditioning by comparing it against a 2D depth map baseline. Despite sharing the same information density derived from a single image, the representation format remains critical for spatial reasoning. While depth maps provide pixel-aligned features, they lack explicit 3D spatial relationships and tend to deviate more from the pretrained behavior of the model. As illustrated in Fig.~\ref{fig:abla}b, using 2D depth maps often results in loss of fine-grained structures. In contrast, by lifting depth into a pose-aware latent space, our method enables the model to directly perceive and process the 3D structure, leading to superior spatial alignment and significantly cleaner geometric details.

\subsubsection{Depth noise augmentation.}
To improve the model's adaptability to imperfect geometric priors from monocular depth estimators, we introduce depth noise augmentation during training. As shown in Fig.~\ref{fig:abla}c, models trained without this augmentation tend to overfit to precise geometric inputs, leading to collapsed surfaces or artifacts when encountering noisy real-world depth maps. In contrast, incorporating stochastic perturbations on point conditions during the training process encourages the model to treat the input as coarse structural guidance rather than an absolute constraint. This augmentation strategy significantly enhances the structural integrity of generated meshes, allowing \method to maintain high fidelity even under sub-optimal input conditions.

\section{Conclusion}
We propose \method, a diffusion model for end-to-end, pose-aligned 3D object generation from single images using a direct 3D latent conditioning mechanism. Our approach completely abandons the canonical space assumption, explicitly locking the generated geometry to the observation space by encoding unprojected partial point clouds directly into the diffusion latent space, thereby fundamentally eliminating pose ambiguity and spatial mismatch. Extensive experiments demonstrate \method's effectiveness in pose-aligned single-object generation and its natural scalability to high-quality compositional scene generation.

\subsubsection{Limitations and broader impact.}
While \method shows promise for pose-aware 3D generation, limitations remain. The accuracy of the spatial alignment inherently relies on the quality of the input partial point cloud, meaning errors from the upstream monocular depth estimator or 2D segmentation masks can propagate to the final geometry. Additionally, heavily occluded regions may still present structural ambiguity during the generation process. Nevertheless, we believe that \method is highly promising and holds significant potential for practical downstream applications such as virtual reality, robotic simulation, and gaming. As a generative method, our model may be misused for the fabrication of deceptive 3D assets or environments, necessitating strong safeguards against misuse.

\bibliographystyle{splncs04}
\bibliography{main}

\newpage
\appendix

\section{More implementation details}
\label{apd:imp-detail}
\subsection{Training data curation}
We construct our training set primarily from the Objaverse~\cite{deitke2023objaverse} and 3D-Front~\cite{fu20213d} datasets. To ensure high geometric quality, we employ a rigorous filtering pipeline. We first apply a quality filter by training a lightweight multi-view classifier to identify and discard objects with poor visual quality, trivial geometry, or multi-object clusters. Subsequently, we perform a strict watertightness check. Because non-watertight meshes severely impede the training of the diffusion VAE, we filter them out by computing the Chamfer distance between the input mesh and its VAE reconstruction, discarding any samples with high reconstruction error. This process yields a refined, high-quality subset of approximately 20k objects.

For each object, we render 24 views with random azimuths and elevations following prior works~\cite{xiang2025structured,meng2025scenegen}. Crucially, the ground truth meshes are rotated to align with the camera look-at direction of the input image to remain consistent with our pose-aware generation objective. 

To robustify the model against partial observations commonly found in complex scenes, we introduce a synthetic occlusion augmentation. We employ a layered rendering pipeline where two objects are rendered simultaneously. When an occlusion occurs, the mask of the foreground object is applied to the background object's depth map. This creates robust training pairs where the input partial point cloud is incomplete, effectively simulating realistic multi-object interaction scenarios.

\subsection{Training Strategy and Experimental Setup}
Migrating a pre-trained diffusion model from a canonical space to an observation space introduces significant training instability. To address this domain shift, we adopt a two-stage curriculum training paradigm.
In the first stage, the model is trained exclusively on the filtered, pure single-object dataset. This phase allows the network to focus entirely on learning the fundamental mapping from the 2D observation space to the unoccluded 3D geometry, establishing a strong baseline for pose-aligned generation.
Once the model demonstrates stable convergence on single objects, we initiate the second stage, continuing the training for an additional 20,000 iterations. In this phase, we seamlessly inject the synthesized occlusion data and more complex scene-level configurations. This progressive transition prevents early training collapse and significantly strengthens the model's capacity to handle the severe occlusions and noisy backgrounds typically encountered in actual multi-object environments.

Another critical challenge in pose-aware diffusion models is shortcut learning, in which the network becomes overly reliant on the provided 3D point cloud and ignores the semantic and textural cues in the input RGB image. While the exact ground-truth point cloud provides excellent spatial anchoring during training, real-world inference relies on depth estimators, which inevitably produce noisy, distorted, or scale-ambiguous point clouds. To bridge this train-test gap and ensure the model benefits from geometric guidance without developing an over-dependency, we inject controlled perturbations into the conditioning data through several distinct mechanisms.

First, we introduce low-probability random scaling to the camera's Field of View during the back-projection of the depth maps. This creates a deliberate, subtle misalignment between the image features and the spatial coordinates of the point cloud, forcing the model to rely more heavily on image features to correct minor geometric distortions. To explicitly simulate inference-time conditions, we also randomly replace the flawless ground-truth point clouds with depth-based point clouds estimated directly by MoGe. Because MoGe predictions contain characteristic noise patterns, such as boundary flying points and local depth inconsistencies. Exposing the model to these artifacts during training dramatically enhances its robustness and generalizability during inference. Moreover, canonical 3D datasets typically assume the camera is perfectly targeted at the geometric center of the object. To break this restrictive assumption, we synthesize off-center data pairs by applying random translations to the camera's look-at target. This simulates scenarios like panning or asymmetric cropping, ensuring the network can generate accurate 3D structures regardless of the object's relative position within the image. In this situation, the input image will still be center-cropped, but the condition point cloud will be backprojected in the original image.

\subsection{Inference Details}

During inference, we use center cropping for the input image and pad it with a 1.15 ratio of the white background, which is consistent with the training phase. We use MoGe to determine the visible point cloud with its estimated depth and intrinsic. The estimated point cloud is then normalized to $[-1, 1]^3$ unit box with calculated constants. After the generation, we could denormalize the generated mesh back to its original place using these constants.

The VAE also needs a normal to encode the shape, and the surface normals are computed directly from the predicted depth map via back-projection. Each pixel is first converted into a 3D point. Local tangent vectors are then estimated from spatial gradients of the 3D point map along the image axes. The normal at each pixel is obtained as the cross product of these two tangents and subsequently normalized to unit length. To improve numerical stability, non-finite depth and normal values are clamped to zero, and degenerate points with near-zero normal magnitude are ignored.

For texture synthesis, we use HunYuan3D-Paint to colorize the generated mesh. Though \method generates in the observation space, the method can still correctly paint the generated shape.

\section{More results}
\begin{figure*}[t]
    \centering
    \includegraphics[width=1.0\linewidth]{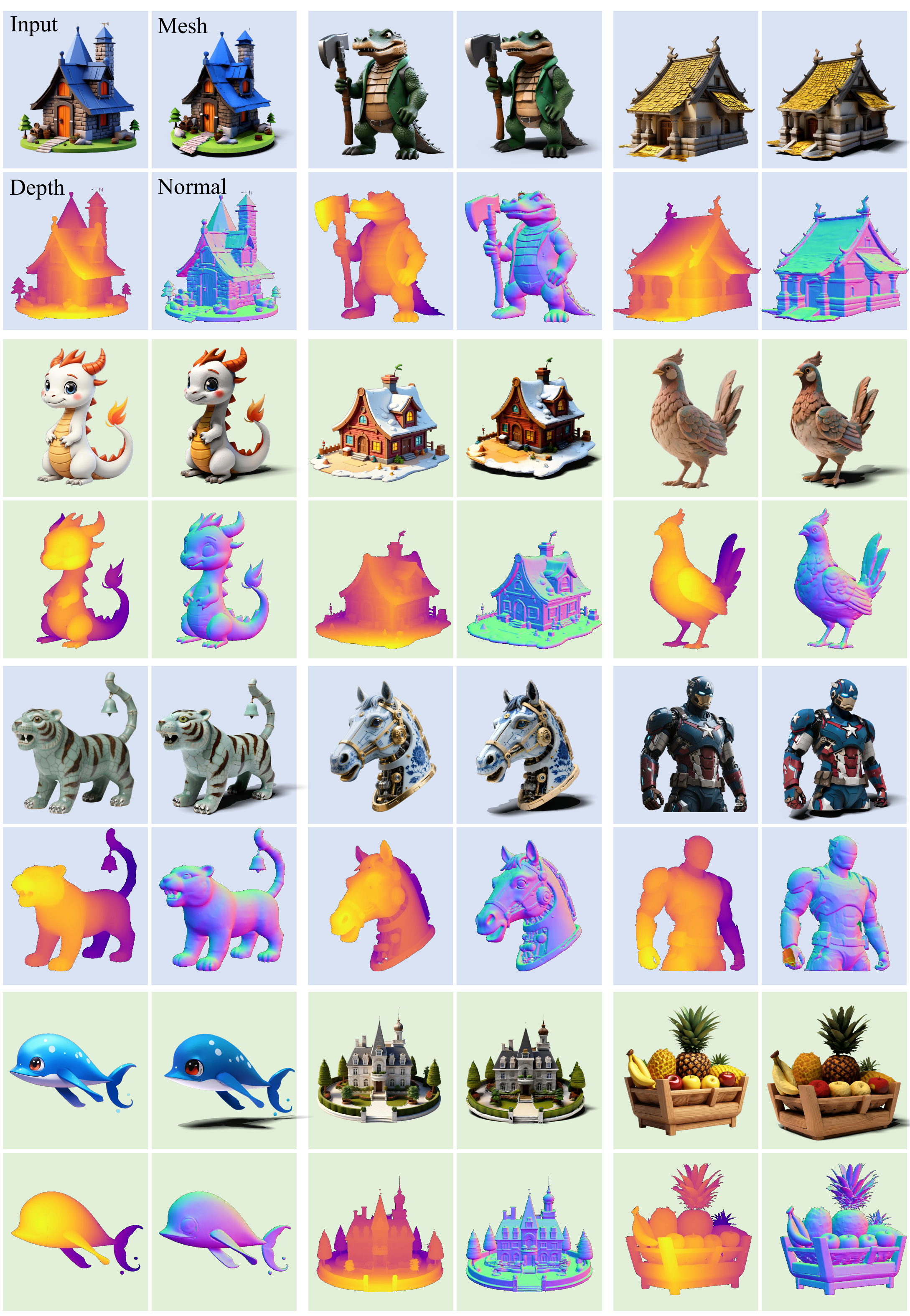} 
    \caption{More results of the generated objects.}  
    \label{fig:appendix1}
\end{figure*}

\begin{figure*}[t]
    \centering
    \includegraphics[width=1.0\linewidth]{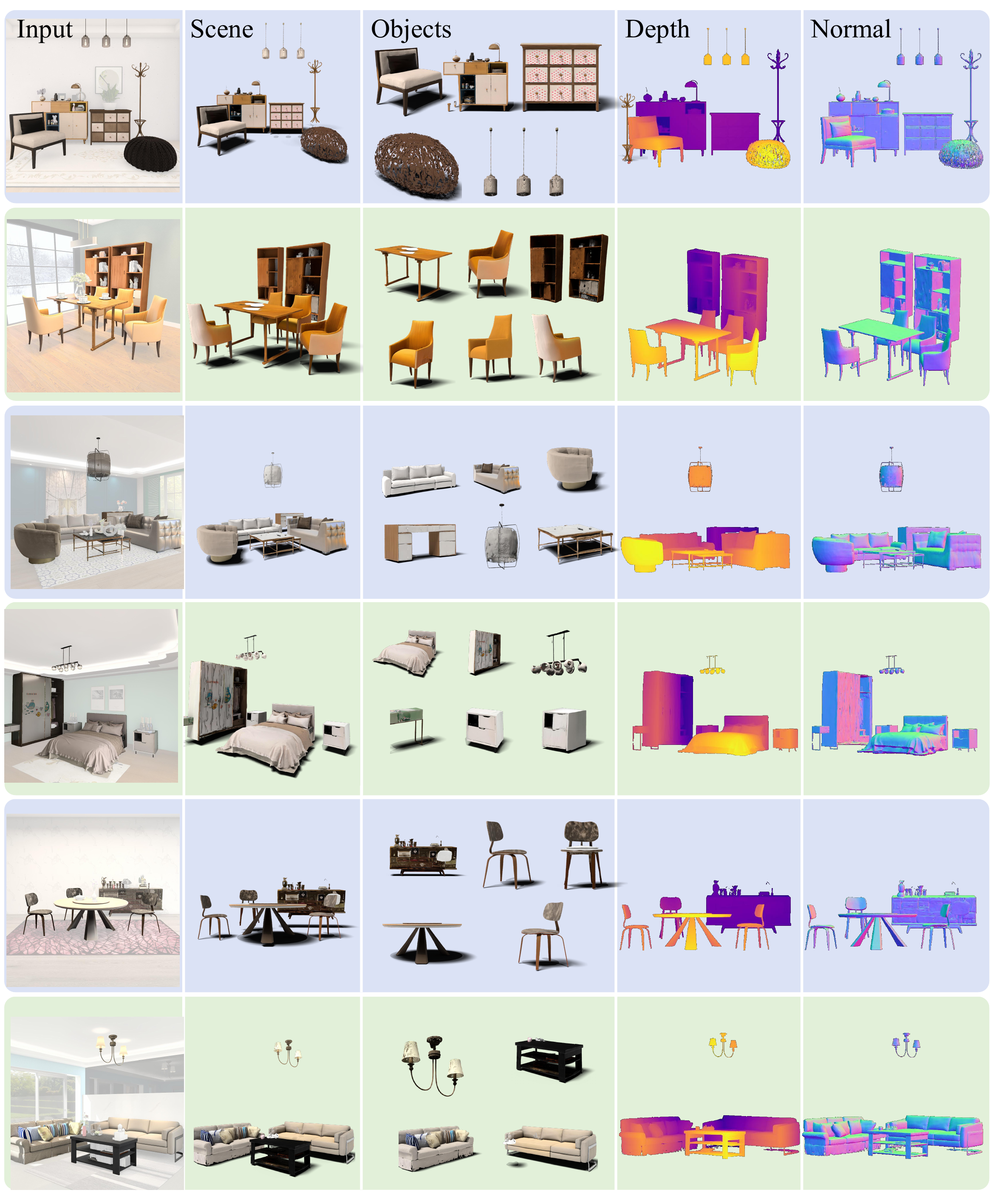} 
    \caption{More results of the generated compositional scenes.}  
    \label{fig:appendix2}
\end{figure*}

We show more results of the generated objects in~\cref{fig:appendix1} and the generated scenes in~\cref{fig:appendix2}.

\end{document}